\documentclass[final]{cvpr}

\usepackage{times}
\usepackage{epsfig}
\usepackage{graphicx}
\usepackage{amsmath}
\usepackage{amssymb}
\usepackage{enumitem}
\usepackage{comment}
\usepackage{color}

\newcommand{\vect}[1]{\mathbf{#1}}

\newcommand{\R}{\mathbb{R}}
\newcommand{\argmax}{\arg\max}
\newcommand{\ip}[2]{\left\langle #1,#2\right\rangle}
\makeatletter
\def\@fnsymbol#1{\ensuremath{\ifcase#1\or \star \or \dagger\or
   \mathsection\or \mathparagraph\or \|\or **\or \dagger\dagger
   \or \ddagger\ddagger \else\@ctrerr\fi}}
\makeatother

\usepackage[pagebackref=true,breaklinks=true,colorlinks,bookmarks=false]{hyperref}
\author{Rami Ben-Ari~\thanks{Equal contribution}~~\thanks{\scriptsize{This work was done while the author was with IBM Research AI}}~~\textsuperscript{1} 
\and Mor Shpigel Nacson \footnotemark[1] ~\thanks{\scriptsize{This work was done while the author was an intern at IBM Research AI}} ~\textsuperscript{2}
\and Ophir Azulai \footnotemark[1] ~\textsuperscript{3}
\and Udi  Barzelay \footnotemark[1] ~\textsuperscript{3}
\and Daniel Rotman \textsuperscript{3} \\
\textsuperscript{1} OriginAI \\
\textsuperscript{2} Technion, Haifa, Israel  \\
\textsuperscript{3} IBM Research AI, Haifa, Israel \\
ramib@originai.co, mor.shpigel@gmail.com \{ophir,udib,danieln\}@il.ibm.com}

\begin{document}

\title{TAEN: Temporal Aware Embedding Network for Few-Shot Action Recognition}

\maketitle

\begin{abstract}
Classification of new class entities requires collecting and annotating hundreds or thousands of samples that is often prohibitively costly. Few-shot learning suggests learning to classify new classes using just a few examples. Only a small number of studies address the challenge of few-shot learning on spatio-temporal patterns such as videos. In this paper, we present the Temporal Aware Embedding Network (TAEN) for few-shot action recognition,  that learns to represent actions, in a metric space as a trajectory, conveying both short term semantics and longer term connectivity between action parts.
We demonstrate the effectiveness of TAEN on two few shot tasks, video classification and temporal action detection and evaluate our method on the Kinetics-400 and on ActivityNet 1.2 few-shot benchmarks. With training of just a few fully connected layers we reach comparable results to prior art on both few shot video classification and temporal detection tasks, while reaching state-of-the-art in certain scenarios.
\end{abstract}

\section{Introduction}
\label{sec:introduction}
Action recognition is one of the fundamental problems in computer vision, with applications such as event detection, clip classification and retrieval in multimedia storage. In this domain, there are two typical tasks, classification and temporal detection. Video classification aims to classify short video clips often {\it trimmed} from a longer footprint, and usually a few seconds long. In temporal action detection setting, actions appear as short temporal sections within a long untrimmed video that can last even for several minutes. Similar to object detection the goal is to detect the time stamp (temporal location) at which a certain action class takes place.

 Deep learning models have been successfully utilized for action recognition \cite{Girdhar17,Hou17,Liu19,W-TALC_ECCV18}. Although these models obtain remarkable results, they require large amounts of labeled training data that often are often unavailable.  
 The task of learning new categories from a small number of labeled examples is known as few-shot learning (FSL). Typical FSL methods are based on meta-learning \cite{MetaLearning_fewShot_2017,MetaLearning_fewShot_TemporalConv_2017}, distance metric learning  \cite{MetricLearning_fewShot_Local_Descriptoy_CVPR_2019,MetricLearning_fewShot_CVPR_2019} and synthesis methods \cite{Synthesis_fewShot_NIPS_2018,Synthesis_fewShot_ICCV_2017}. While few-shot learning has been extensively studied in the context of visual recognition \cite{Hariharan_FewShotImageClassif17,Kwonjoon_FewShotImageClassif19,Snell_FewShotImageClassif17,Sung_FewShotImageClassif18}, few studies address the challenge of learning from a few instances to detect spatio-temporal patterns. In particular, only a few works have been proposed for few-shot learning in action recognition and specifically for video classification \cite{TARN_Bishay19,ProtoGAN19,TemporalAlignment,FSL_ECCVW2020,Linchao18} and temporal action detection \cite{Yang18}. To alleviate the annotation labour,  weakly supervised strategies \cite{Weakly_ActionLocalization_Nguyen_CVPR18,W-TALC_ECCV18} are suggested, where tagging is conducted just over the entire video. However, in cases where new classes may appear on the fly or in rare class types, few-shot learning methods has no alternative.  In FSL one can learn {\it new categories} from just a few video examples, often one to five. This is in contrast to transfer learning where tens or hundreds of new labeled examples are needed to learn a new class.
  In this paper, we address the task of few-shot learning video classification and show also an extension for temporal action detection.

 
 \begin{figure*}[t!]
    \centering
     \includegraphics[width = 0.95\linewidth, trim=0cm 7cm 0.2cm 6.6cm, clip]{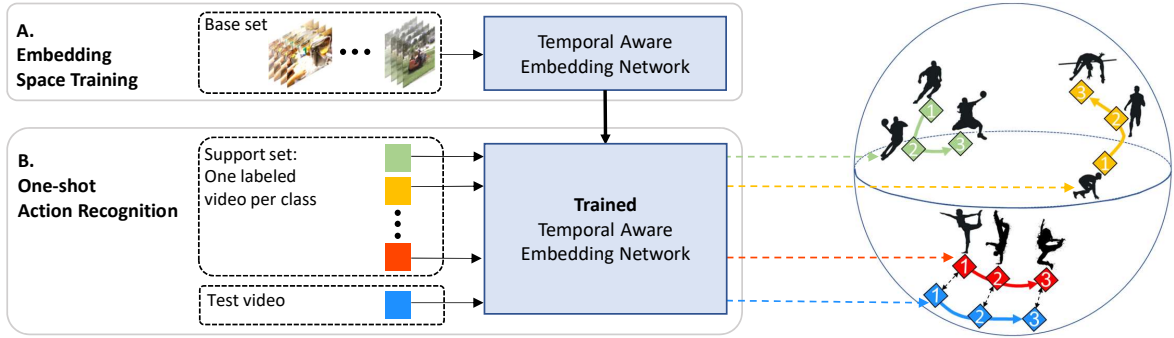}
    \caption{\textbf{Action representation as a trajectory in a metric space.} We train an embedding space where actions are represented as trajectories passing through sub-actions. Jointly learning the sub-action prototypes and the embedding function allows our representation to be optimized for both short and long term action granularity. Using the trained embedding network, we embed $n$ videos from different classes ($n$-way, $1$-shot scenario) and the query test video. The query is then classified based on distance between the associated trajectories illustrated by the black arrows between red and blue trajectories.}
    \label{fig: main idea}
\end{figure*}

Action recognition in videos has been greatly advanced thanks to powerful features such as dense trajectories \cite{Dense-Trajectories}, and deep features \eg  the Two-Stream networks \cite{Two-Stream-Networks}, C3D \cite{C3D}, and I3D \cite{I3D}. Prior to deep features this was tackled by extracting spatio-temporal local descriptors from space-time interest points \cite{Dense-Trajectories} (iDTF) presenting trajectories in pixel-time space. The sparse patterns of trajectories were then analyzed to distinguish between different actions. However, these trajectories do not convey the semantic information gained in deep representations of two-stream networks, I3D or C3D. Yet, common deep features often miss the long term dependencies involved with complex actions, by processing short video segments (usually 16 consecutive frames). 

In this work, we propose a novel approach based on Distance Metric Learning (DML) for few-shot action recognition, called Temporal Aware Embedding Network - TAEN, as described schematically in Fig. \ref{fig: main idea}. We start with decomposition of the video into {\it ordered} temporal segments, called {\it sub-actions}. In contrast to the whole action that can last several seconds, sub-action is only part of the action in the timeline. We consider each action as a consecutive set of sub-actions and train an embedding function that can represent an action as a trajectory in a metric space, considering also the long term sub-action dependencies. This representation preserves the temporal order of the sub-actions while carrying the semantics encoded in deep features. Using a trajectory distance enables our method to distinguish between even fine-grained actions, manifested as similar actions that differ only in some intermediate part (i.e a sub-action).

We evaluate TAEN on Kinetics-400 data set \cite{Kinetics17}. For extension of TAEN to temporal action detection we build upon the BMN temporal region proposal \cite{lin2019bmn} and evaluate our method on a benchmark previously suggested by \cite{Yang18} on ActivityNet 1.2 \cite{ActivityNet1_2_CVPR15}. Our method is computationally cheap and scalable, relying on training of only a few fully connected layers for learning the embedding function and the sub-action prototypes. TAEN allows classification of new action categories fairly well, even with a single example.

Our contributions are: \textbf{(1)}, we suggest a new metric learning method that encodes the long term sub-action connectivity into an embedding space, eventually representing the actions as trajectories in a {\it metric space}. \textbf{(2)}, Our model jointly optimizes for sub-action prototypes and the embedding function. \textbf{(3)}  
We suggest a novel loss function that allows an effective learning of class action trajectories. \textbf{(4)}, using the same model we present comparable results to prior art in video classification and temporal action detection.

\section{Related work}
\subsection{Sub-actions} 
Decomposing actions into characteristic sub-actions has been studied before \cite{Hou17,Liu19}. A common practice in these works consists of aggregating and pooling local features from sub-action segments. While naive approaches use mean or max pooling, recent studies extend the pooling techniques by incorporating them into Deep Neural Network (DNN), namely NetFV \cite{Lev16} and NetVLAD \cite{Girdhar17}. By looking for correlations between a set of primitive action representations, ActionVLAD \cite{Girdhar17} has shown state-of-the-art performance in several action recognition benchmarks. However, cluster-and-aggregate based methods such as NetVLAD assign soft-clusters to every frame in the video and therefore also use a too fine granularity, particularly for complex actions. These methods further ignore the temporal ordering that might be crucial for recognizing an action from a single example. Therefore, these methods are not well suited for the few-shot action recognition settings. 

Typically a few sub-actions are enough to represent simple actions, while others would need to be represented with more parts. Based on this assumption, Hou \etal \cite{Hou17} suggested a temporal action detection method, trying to optimize the number of sub-actions for each action class. However, this method uses the hand crafted dense trajectory features (iDTF) \cite{Dense-Trajectories} and does not handle the few-shot setting.
In \cite{Liu19}, the sub-action notion is used in the context of weakly supervised temporal action detection.

\subsection{Few-shot video classification}
Zhu and Yang \cite{Linchao18} suggest a method based on memory networks and meta-learning that require heavy computation and space resources to learn new representations for each episodic task. They further use a single embedding vector to represent the entire video, that overlooks the temporal structure of the action and focuses mainly on the most distinguishable segment of a video. The approach by Bishay \etal \cite{TARN_Bishay19} named TARN, addresses FSL on short video segments by calculating the relation between a query and support video  with a similarity measure between {\it aligned} segments. Their model searches for relations between short-length segments (16 frames) and therefore still relies on an extremely fine action granularity, while in the relation process the temporal order is further lost. Recently \cite{TemporalAlignment} showed SOTA results in few shot video classification by action decomposition and temporal alignment using Dynamic Time Warping (DTW). We argue that our method is more efficient by finding the optimal prototypes without the DTW alignment which is $\mathcal{O}(N^2)$. Optimized prototypes yet suggest robustness to slight miss-alignment in the video clips. 
\subsection{Few-shot temporal action detection}
While metric learning approaches have found their way to few-shot learning object detection tasks \cite{karlinsky2019repmet,FewShotObjectDetection_ICCV19}, their use in the few-shot temporal action detection task has been overlooked. The recent attempt taken for this task by Yang \etal \cite{Yang18} suggests a  meta-learning method based on Matching Networks and uses a Long-Short Term Memory (LSTM) video encoder. Their method is computationally heavy as it requires optimization over many different episodes. Moreover, to tackle action detection, they use a  “sliding window” approach for action proposals, which further sets a high computational cost and a high imbalance in the training set, resulting high false-positive rates and inflexible activity boundaries. Recent work shows that a more effective and efficient way is using a temporal region proposal method \cite{Strongly_ActionLocalization_FRCNN_CVPR18,R_C3D_Network_Action_Detection_ICCV17,Strongly_ActionLocalization_SSN_ICCV17}. In fact, new temporal region proposal methods such as \cite{lin2019bmn,BSN_ECCV2018} are now suggested as part of temporal action detection pipelines. \cite{RC3D-arXiv2018} addresses the problem of effective temporal region proposal for improved detection accuracy.

In this paper we learn to present actions as {\it ordered} temporal segments, namely {\it sub-actions}. In contrast to the whole action that can last several seconds, sub-action is only part of the action in the timeline. We consider each action as a consecutive set of sub-actions using sub-actions semantics as well as their long term dependencies for imoproved action recognition. Our method builds on {\it jointly} learning the prototypical sub-actions and the embedding function keeping their temporal order, eventually describing actions as a {\it trajectory} in a metric space. Actions are commonly separated in the embedding space by the semantics of objects in the frame and the background. We define an action trajectory as a parametrized curve in an embedding space, discretized by sub-actions. Each action signature is then obtained by a {\it temporally ordered} set of points (prototypes of sub-actions) representing different parts of an action in the embedding space. This type of representation allows to better discriminate between similar actions even when they take place in the same scene. The unique trajectory associated with each action allows recognition of a new action class from only a few examples. Our model requires training of just a few FC layers, making it efficient in both training and inference. Our suggested method called Temporal Aware Embedding Network - TAEN learns to represent videos as trajectories in the feature space.

We evaluate TAEN on Kinetics-400 data set \cite{Kinetics17}. For extension of TAEN to temporal action detection we build upon the BMN temporal region proposal \cite{lin2019bmn} and evaluate our method on a benchmark previously suggested by \cite{Yang18} on ActivityNet 1.2 \cite{ActivityNet1_2_CVPR15}.


\section{TAEN}
\label{sec:Proposed method}
Fig. \ref{fig: model_architecture} presents our model architecture. We define the {\it trajectory} in the DML space as a collection of sub-action prototypes with temporal order. Actions are then represented in the metric space by a set of ordered prototypes. In the few-shot video classification and temporal action detection, our goal is to train a model that will be able to generalize to new unseen classes. To this end, we propose a novel method that relies on two main steps:
\begin{enumerate}
    \item Learn an embedding space where actions are represented by well-separated trajectories. Action trajectories are represented by $a \in \mathbb{N}$ temporally ordered centers, one for each sub-action.
    \item Learn jointly the prototypes and the embedding function.
    \item Classify new videos according to their {\it trajectory} signature.
\end{enumerate}

\subsection{Embedding architecture}

\begin{figure*}[ht]
\centering
    \includegraphics[width=1\linewidth]{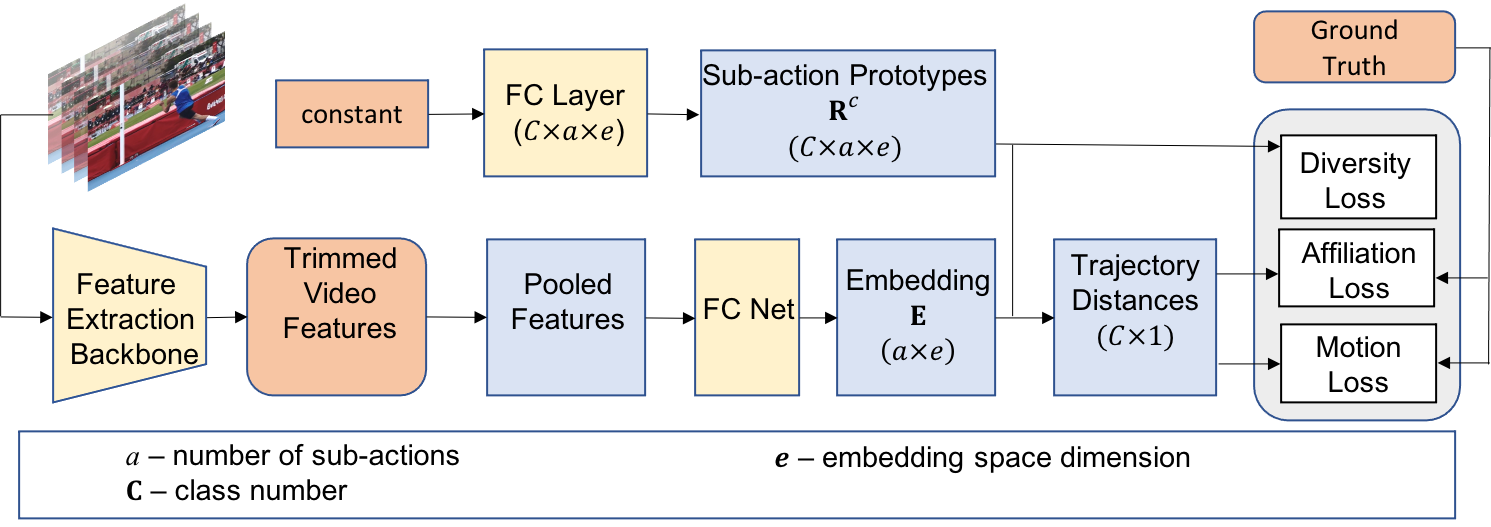}
    
    \caption{\textbf{The TAEN model} is designed to jointly train the DML-embedding (lower branch) along with the $a$ sub-action prototypes (upper branch).}
    \label{fig: model_architecture}
\end{figure*}

Our network architecture is inspired by the RepMet architecture \cite{karlinsky2019repmet} that enables the parallel training of the embedding function and the prototypes. This architecture allows joint learning of multiple prototypes per class, sub-actions in our case, while training them with the embedding to reach optimized embeddings and prototypes.

We design a novel loss function to allow discrimination of trajectories in the embedding space. Our training is done in batches but for simplicity we will describe a single video flow through the architecture. The model architecture is depicted in Fig. \ref{fig: model_architecture}.

The model input is a series of video features $\vect{X}\in\R^{T\times d_{feat}}$ computed using some pretrained backbone (\eg C3D or I3D architectures), where $T$ is the number of video segments, namely sub-actions and $d_{feat}$ is the dimension of the backbone output features. We divide the video into $a$ segments and calculate the representation for each segment by average pooling: $\vect{X}_{p}\in\R^{a\times d_{feat}}$. Here, $a$ represents the number of sub-actions. Next, we use $\vect{X}_{p}$ as an input to the DML network, which consists of a few FC layers with RELU activation functions. We denote the embedding network output for certain class $c$ as $\vect{E}\in\R^{a\times e}$, 
where $e$ is the embedding space feature dimension. Typically, $e \le d_{feat}$.

\subsection{Sub-action prototypes}
In order to allow joint training of multiple sub-actions, we build on an architecture that adds a secondary branch as described in Fig. \ref{fig: model_architecture}. This branch allows computing our sub-action prototypes for each action class. One can initialize the prototypes per-class by K-means clustering. Note that at this point video segments are represented by pre-computed deep features. However we choose to initialize the sub-action prototypes by using a constant scalar as input to a FC layer that yields $C\times a\times e$ parameters, where $C$ is the number of action classes in the training set \cite{karlinsky2019repmet}. This is equivalent to random seed clustering initialization. We reshape the fully connected network output into $C$ segments (one for each class) denoted by $\vect{R}^c\in\R^{a\times e}, \forall c=1,...,C$. We also denote $\vect{R}^c_i\in\R^{e}$ as the representation of the $i$-th sub-action of class $c$ in the DML-embedding space. Note that $\vect{R}^c_i$ is the $i^{\text{th}}$ dimension of tensor representation of $\vect{R}^c$.

\subsection{Loss function}
From the two branches described above, we obtain the video embedding in the DML-space $\vect{E}$ and the prototypes for each class $\vect{R}^c$. These representative centers denote the learned sub-actions for our trajectories in the embedding space. Using the video embedding and the computed prototypes, we can calculate the distance associated with the trajectories in the embedding space as: 
\begin{equation}\label{eq: trajectory distance}
d\left(\vect{E},\vect{R}^c\right) = \frac{1}{a}\sum_{i=1}^a d\left(\vect{E}_i, \vect{R}^c_i\right), \forall c= 1,...,C
\end{equation}
where $\vect{E}_i$ is the $i^{\text{th}}$ sub-action of $\vect{E}$ and $d\left(\cdot,\cdot\right)$ is some vector distance metric e.g., euclidean or cosine distance. 
As the distance metric $d$ we use the cosine distance over a hypersphere, i.e., unit normalized embedding vectors $\vect{E}$, $\vect{R}^c$ (see in Fig. \ref{fig: main idea}).
Using the prototype representations and the trajectory distance metric, we can calculate the training loss for our model. The loss consists of the following three components:
\begin{enumerate}[leftmargin=*]
    \item \textbf{Affiliation loss:} For a given embedding video $\vect{E}$, this loss minimizes the distance of the embedding sub-actions and their prototypes (jointly learned):
    \begin{equation}
        \mathcal{L}_{aff} = \sum_{i=1}^a d\left(\vect{E}^c_i,\vect{R}^c_i\right)\,,
    \end{equation}
    where $c$ is the true class index (extracted from labeled data) and $\vect{R}^c_i$ denotes the representative center of the $i^{\text{th}}$ sub-action of class $c$.
    \item \textbf{Motion loss:}
    This loss measures the deviation in the trajectory gradient, approximated by the vector of change between consecutive sub-actions:
    \begin{equation}
        \mathcal{L}_{mot} = \sum_{i=1}^{a-1} -\langle{\vect{E}^c_{i+1}-\vect{E}^c_{i},\vect{R}^c_{i+1}-\vect{R}^c_{i}\rangle},
    \end{equation}
    where $\ip{\cdot}{\cdot}$ denotes the inner product operator. Note that this loss drives the embedding space to higher diversity between sub-actions due to dependency on the norm of vectors between sub-actions. It further drives the space toward alignment between the model and test motion vectors.
     
    \item \textbf{Diversity loss:} This loss term aims to prevent the representation of sub-action prototypes, from collapsing into one point in the embedding space. Since short trajectories determined by close points are less discriminative, we enforce the diversity loss on the corresponding sub-actions. Semantically, this loss ensures that the sub-action prototypes in each class, will be sufficiently apart from each other. This is achieved by penalizing large correlations between different sub-action prototypes:
    \begin{equation}
    \mathcal{L}_{div} = \sum_{c=1}^C \sum_{i=1}^a \sum_{\substack{j=1 \\ j \neq i}}^a 1-d\left(\vect{R}^c_{i},\vect{R}^c_{j}\right)
\end{equation}
   
\end{enumerate}
Finally, the total loss consists of the weighted sum of the above three terms:
\begin{equation} \label{eq: total loss}
    \mathcal{L}_{total} =  w_{aff}\cdot \mathcal{L}_{aff} + w_{mot} \cdot \mathcal{L}_{mot} + w_{div} \cdot \mathcal{L}_{div}\,,
\end{equation}
where $w_{aff},w_{mot}$ and $w_{div}$ are tuned as part of the hyper-parameters tuning process.
In the next section we provide additional details on our method implementation for action classification and for temporal action detection.

At test time we measure similarity between actions as a weighted distance between discrete point-wise similarity and motion:
\begin{equation}
    d_{test} = 
    \sum_{i=1}^a w_{aff} ~ d\left(\vect{E}_i,\vect{R}^c_i\right)\ +w_{mot}\langle{\vect{E}_{i+1}-\vect{E}_{i},\vect{R}^c_{i+1}-\vect{R}^c_{i}\rangle}
    \label{eq:test_dist}
\end{equation}
\section{Evaluation}
In the few-shot scenario, our goal is to train a network that generalizes well to new action classes. In this setup, we are given a training set that consists of labeled videos from different classes. This set is denoted as the "base set" and is only used during training. For testing, we are given a small number of videos from {\it new} classes that were not available in the base set. Our task is video classification in trimmed videos or temporal action detection in untrimmed videos.
\subsection{Training}
For the classification task, we follow the experimental set-up in previous works using the same backbone feature extractor-C3D pretrained on Sports-1M \cite{Sports-1M_CVPR14}. Using the pre-trained backbone on an external dataset is justified to allow fair comparison to previous works \cite{TARN_Bishay19,Linchao18}. For the temporal action detection experiments we pretrain our I3D backbone only on the {\it base} set of ActivityNet 1.2 to avoid any chance of feature "contamination". 
We follow the evaluation process as in \cite{Yang18} detailed in section \ref{sec: Experimental results}.

\subsection{Testing}
In the few-shot setting, tests are determined by episodes of $n$-way, $k$-shot tasks, where at each test episode we sample $k$ videos from $n$ different classes (total of $n \times k$ videos) to build the support set. In our model, each episode is represented by $n$ trajectories (one representative trajectory per-class). Representation over k-shots is computed by average pooling over corresponding sub-actions. 

\begin{figure*}[ht]
\centering
    \includegraphics[width=1\linewidth]{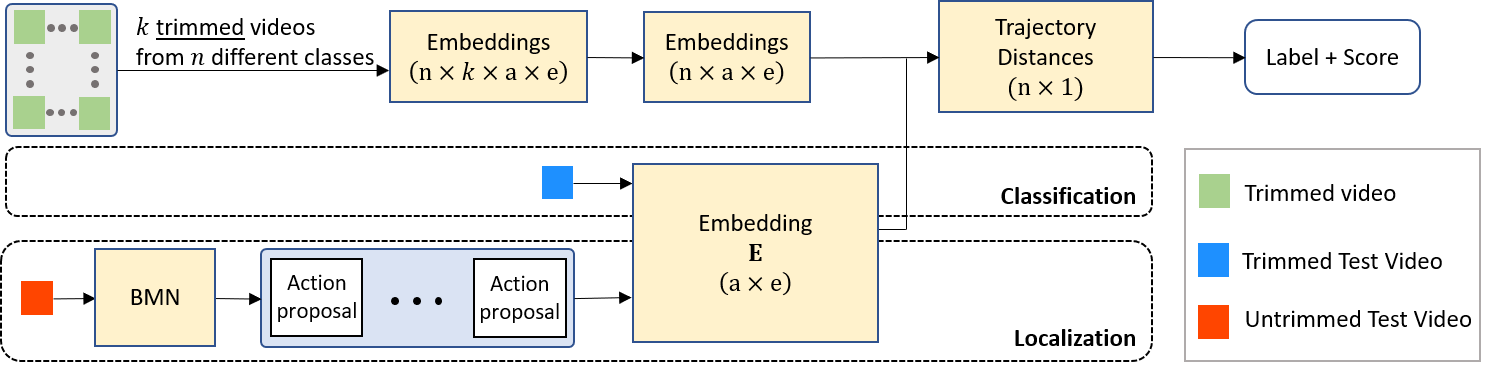}
    \caption{\textbf{Test Model Architecture:} \underline{Top branch:} Using the same flow as in Fig. \ref{fig: model_architecture} and the trained DML, we obtain the embedding for all support set videos and the corresponding sub-actions. \underline{Middle branch - classification:} For the classification task, given a query trimmed test video, the video is decomposed to $a$ sub-actions and mapped into our DML space as a set of ordered points creating a trajectory. Classification is then performed based on the trajectory distance from the support classes using minimal distance. \underline{Lower branch - Detection (Localization):} For the temporal detection task we are given one untrimmed video. We use the BMN temporal region proposal network to decompose the video into action proposals. Proposals are then treated as trimmed videos and classified in the support set. We use the trajectory distances to compute a confidence score. Rejecting background segments with low confidence score and Non-Maximal Suppression yields our final temporal temporal detection.} 
    \label{fig: Test}
\end{figure*}

For the video classification task, each trimmed video is mapped to a trajectory in the embedding space and the class is derived using the nearest trajectory (see Eq.~\eqref{eq:test_dist}).
For the task of action detection, we build upon a {\it temporal} region proposal network \cite{lin2019bmn} and follow the same process with the candidate proposals. We then filter out background segments by thresholding over the scores. The test pipeline is described in Fig. \ref{fig: Test}.

\textbf{TAEN based video classification:} Using the trained embedding network, we build the support set trajectories by computing sub-action prototypes for the $n$ classes. 
Next, given a query video, belonging to one of the $n$ classes, we calculate the video embedding using the trained model. Classification is then based on the nearest trajectory using the trajectory distance metric in Eq.~\eqref{eq:test_dist}.

\textbf{TAEN based temporal detection:} Similarly to the classification task, the base and the support set are trimmed videos that are defined by the ground truth annotations in untrimmed videos. The trajectories for each support class are calculated in the same way as in the classification task. Yet in the detection task, the query is an untrimmed video. We therefore decompose the video into temporal regions using a standard temporal action proposal method. Proposal segments are then classified and scored according to trajectory distance. 
The probability of an action proposal belonging to certain action class \ie one of the $n$-way classes is set by:
\begin{equation}
p_m^c = \exp\left(-\frac{d^2\left(\vect{E},\vect{R}^c\right)}{2\sigma^2}\right)\,,
\end{equation}
where $\sigma$ is an hyper-parameter that controls the standard deviation of the probability measure.
Then, using the probabilities we associate the $m^{\text{th}}$ action proposal to one of the classes $\hat{c}_m$ and calculate the score:
\begin{equation}
    \hat{c}_m = \argmax_c p_m^c, ~s_i = p_m^{\hat{c}_m}   \label{eq: TAEN confidence score}
\end{equation}
where $s$ is the score of the $m^{\text{th}}$ proposal. Eventually, background segments are rejected based on low confidence scores.

\section{Experimental results} \label{sec: Experimental results}
\textbf{Implementation details:} To train the proposed architecture we use SGD. We set the batch size $B=30$, learning rate $\eta=5\cdot10^{-4}$ and momentum $m=0.9$ to minimize the loss defined in Eq.~\eqref{eq: total loss}, in the video classification task. We used $\alpha=0.2$ as the margin for the sub-action loss.
In the classification experiments we obtained the best results with $w_{aff}=1,w_{mot}=0.5$ and $w_{div}=20$. In the detection experiments we used the same parameters except a larger batch size of $B=50$. In addition, we used $\sigma=0.5$ as the probability standard deviation hyper-parameter. We obtained best performance for $a=5$, i.e. decomposing the videos into 5 sub-actions.
\subsection{Classification}
We start by evaluating our architecture performance on the classification task. In \cite{Linchao18}, the authors introduced a dataset for few-shot classification which is a modification of the original Kinetics-400 dataset \cite{kay2017kinetics}, consisting of 400 categories and 306,245 videos, covering videos from a wide range of actions and events, e.g., “dribbling basketball”, “robot dancing”, “shaking hands” and “playing violin”.  The modified dataset contains videos from 100 categories out of 400 available ones randomly selected from the original Kinetics-400 dataset. Each class category contains 100 videos. In addition, the authors divided the dataset into 64, 12, and 24 non over-lapping classes for training, validation and testing respectively. In our classification evaluation, we use the same dataset and evaluation protocol as defined in \cite{Linchao18} and followed by \cite{TARN_Bishay19}.

\textbf{Feature extraction} We follow the feature extraction protocol of C3D trained on Sports-1M \cite{Sports-1M_CVPR14}, as in \cite{TARN_Bishay19,Linchao18} to perform a fair comparison to the previous works. Note that the feature extraction network is trained on a different dataset than Kinetics. Using the pre-trained C3D architecture, the 4096D features are extracted from the the last FC layer (i.e. FC7) of the network, corresponding to 16 consecutive frames, and then used as input for our architecture as illustrated in Fig. \ref{fig: model_architecture}. We divide the video into "$a$" non-overlapping segments and use average pooling to obtain the representation for each segment. Note that each video segment represents a different sub-action in the original video. $\vect{X}_p$ is used as an input to our embedding network. In these experiments we used a network with two hidden layers and RELU activation function to obtain the embedding network output $\vect{E}\in\R^{a \times e}$ with $e=2048$.

\textbf{Evaluation protocol}
We compare our model with several baselines and three previous methods. For each class, we randomly choose $k$ videos where $k=1,...,5$ is the number of shots. We use these $n\times k$ videos to calculate class trajectories as described in Section \ref{sec:Proposed method}. Then, we randomly draw one additional test video belonging to one of the $n$ classes, and associate the video embedding to one of the $n$ classes based on the lowest trajectory distance (Eq.~\eqref{eq:test_dist}).  We repeat this episode for 20,000 iterations and evaluate our performance based on the average accuracy.

\textbf{Results:} Table \ref{table:trimmed results} shows our results compared to previous work on the Kinetics few-shot dataset and under the same protocol, as well as a baseline and an ablation study. As our baseline we use the C3D feature space without our embedding, which represents the classification accuracy when each video is represented by a single C3D feature using max-pooling. Without our metric learning C3D shows inferior results (see Table \ref{table:trimmed results}). Our model benefit is manifested when actions are represented by their trajectories in the embedding space. The gaps with respect to C3D of absolute 6.85 points for 1-shot and 4.22 points for 5-shots further show the benefit of our method. 
Our model also outperforms previous methods of \cite{TARN_Bishay19} and \cite{Linchao18} while being inferior with respect to temporal alignment based method of \cite{TemporalAlignment}. The results from our method present a growing gap for higher number of shots, with respect to previous methods of \cite{TARN_Bishay19,Linchao18} as adding more samples (shots) allows smoother and more reliable trajectories. The ablation study shows the impact of each term in our loss function, showing that the contribution of diversity and motion terms is more significant for higher shots. The break down with vanishing affiliation weight is expected since it directly associates the actions with their relevant class.

\begin{table*}[ht]
\begin{center}
\begin{tabular}{l c c c c c | c c c c}
\hline
\multicolumn{6} {c} {Benchmark} & \multicolumn{3} {c} {Ablation} \\
Method &  C3D$^1$ & CMN \cite{Linchao18}& TARN \cite{TARN_Bishay19}& TAM \cite{TemporalAlignment}& TAEN &  $w_{aff}=0$ & $w_{div=0}$ & $w_{mot}=0$ \\
\hline
1-shot & 60.42 & 60.50 & 66.50 & 73.0 & 67.27 &  24.65 & 67.16 & 67.61 \\
2-shot & 70.10 &  70.00 & 74.56 & - & 74.87  & 25.30 & 74.15 & 74.19 \\
3-shot & 75.09 &  75.60 & 77.33 & - & 79.06  & 24.70 & 77.97 & 78.32 \\
4-shot & 77.80 & 77.30 & 78.89 & -  & 81.78  &  24.52 & 80.01 & 80.55 \\
5-shot & 79.76 &  78.90 & 80.66 & 85.8 & 83.12 & 24.79 & 81.04 & 81.55 \\
\hline
\end{tabular}
\end{center}
\caption{Benchmark and ablation study: Video classification accuracy for different shots, compared to state of the art. Our method outperforms the C3D baseline and the prior art of \cite{TARN_Bishay19,Linchao18}. Yet, it is inferior to \cite{TemporalAlignment}, that performs alignment between action segments and associated with $\mathcal{O}(N^2)$ computational cost. C3D means directly using the backbone features in test, without using DML. Ablation: ($w_.=0$) shows the impact for different parts of the loss function, with the affiliation loss as the main part. All of our tests were conducted with the same random seed.}
\label{table:trimmed results}
\end{table*}

\begin{figure}
    \centering
    \includegraphics[width = 0.8\linewidth]{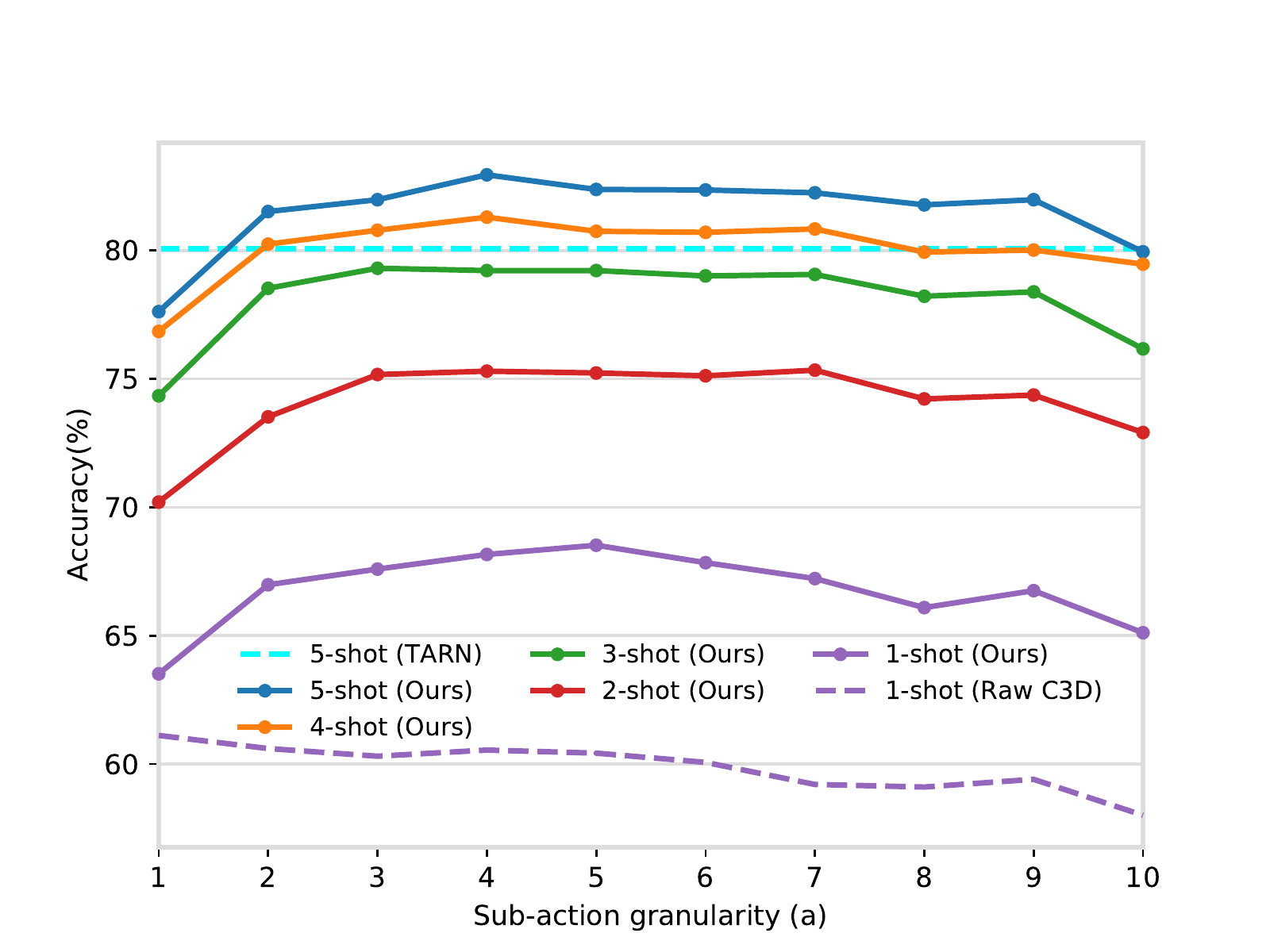}
    \caption{Kinetics 5-way Accuracy vs. number of sub actions. The dashed cyan line indicates the results from \cite{TARN_Bishay19} for 5-shots.}
    \label{fig: subactios_kinetics}
\end{figure}

Next, we depict in Fig. \ref{fig: subactios_kinetics} the variation of accuracy with respect to sub-action granularity and number of shots. We observe that 3-5 sub-actions are sufficient to capture the action structure and enable improved recognition capability. We further show for comparison the performance of \cite{TARN_Bishay19} for 5-shots (dashed cyan) as reference. An interesting comparison is to the raw C3D features, \ie without using our embedding network. Fig. \ref{fig: subactios_kinetics} shows a large margin in 1-shot compared to raw C3D features (see bold and dashed purple lines in Fig. \ref{fig: subactios_kinetics}). The gap between TAEN and raw C3D $1$-shot results is maintained for any number of sub-actions, emphasizing the contribution of our embedding model in the $1$-shot scenario.

\begin{figure}
    \centering
    \includegraphics[width = 0.8\linewidth]{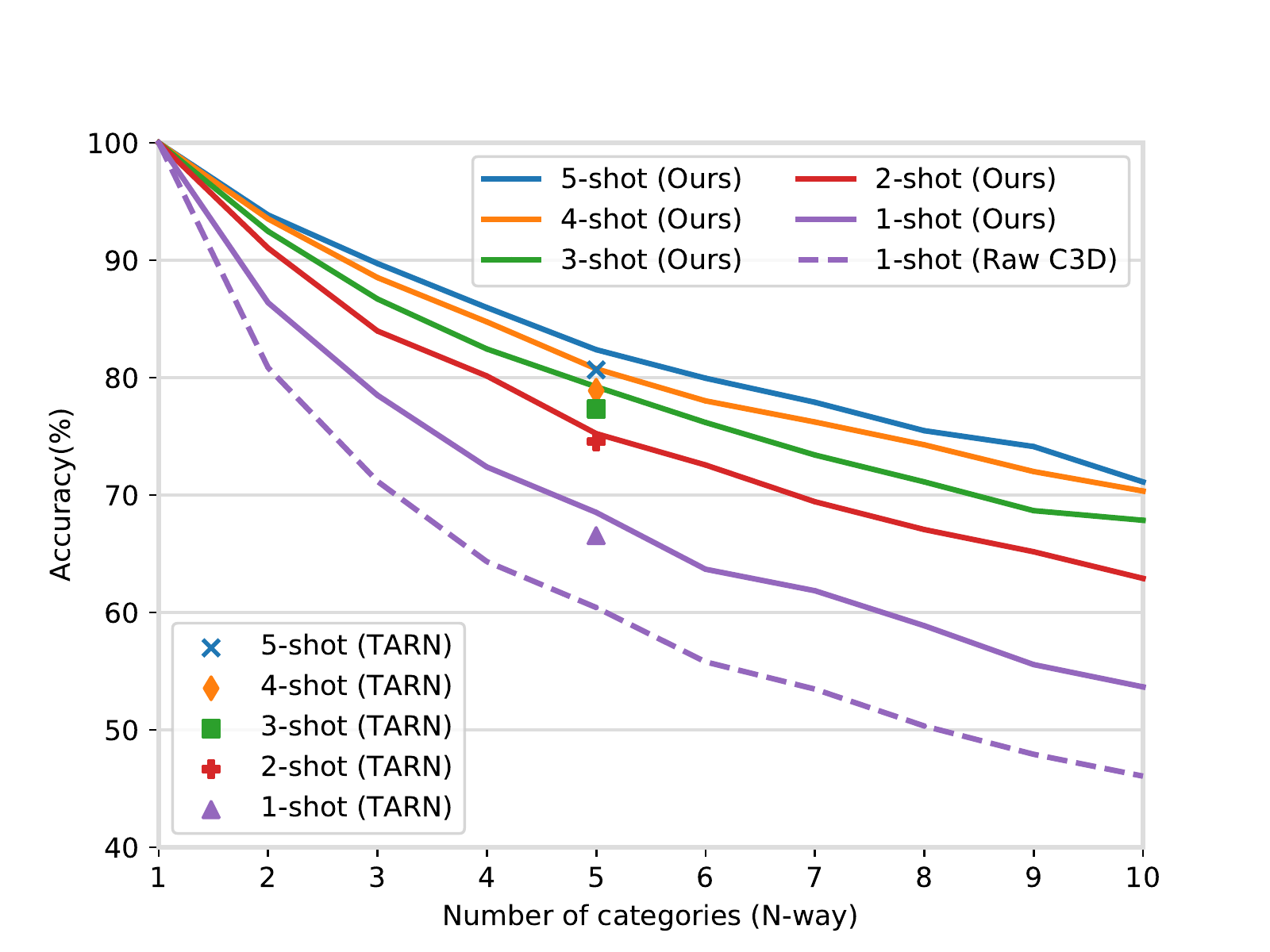}
    \caption{Accuracy variation with number of shots and categories (n-ways) with comparison to baseline (without embedding). Shaped points show the results from \cite{TARN_Bishay19}.}
    \label{fig: accuracy vs. n-ways}
\end{figure}

For detailed analysis, we show in Fig. \ref{fig: accuracy vs. n-ways} the variation in accuracy of our model, with respect to number of categories $n$ (n-ways). As the number of categories grows, the chance for error is raised, causing the decrease in accuracy. In fact, this is a typical scenario in few-shot benchmarks. The figure also shows the sharp decrease in accuracy with $n$ for low number of shots (1,2), while for higher shots (4,5) there is a moderate degradation. This indicates the need for over 4-5 samples to learn a more reliable trajectory for an action class. The dashed purple line shows the lower bound obtained from raw C3D features. Note again the higher impact of our model especially in low shots.
\subsection{Temporal Detection}
In this section we illustrate an extension of our model to the temporal few-shot action detection task. To this end, we evaluate our model on the ActivityNet1.2 dataset \cite{ActivityNet1_2_CVPR15}. This dataset contains roughly 10k untrimmed videos with 100 activity classes. The 100 activity classes are randomly split into 80 classes (ActivityNet1.2-train-80) for training and 20 classes (ActivityNet1.2-test-20) for testing. Our few-shot action detection network is trained on videos containing only the 80 classes in the training set, denoted by ActivityNet-train-80, and is tested on the other 20 classes in the validation set, denoted by ActivityNet-val-20, following the protocol in \cite{Yang18}.

\textbf{Evaluation: }  In order to fit this task to our model we use the Boundary Matching Network (BMN), a temporal region proposal method \cite{lin2019bmn}. We train the BMN on ActivityNet1.2-train-80 for a class-agnostic temporal region proposal. Note that our temporal region proposal is also trained only on the base set to avoid any train-test contamination. For each test video, BMN outputs 100 proposals represented by the start and end time of the actions and associated with a confidence score. We use this score for eliminating some proposals with a score threshold below 0.2. The remaining regions are then processed by TAEN in the same way as trimmed videos. We then reject background segments based on low TAEN confidence scores obtained from \eqref{eq: TAEN confidence score}, and use standard Non-maximum Suppression (NMS) for final temporal detection. We show results on 
I3D trained {\it only on the base set}. 

For evaluation we follow the typical protocol in few shot object detection task \cite{karlinsky2019repmet}. Our benchmark contains multiple random episodes (instances of the few-shot detection tasks). We randomly choose $n=5$ test classes in ActivityNet1.2-test-20. For each test class we randomly sample $k$ videos (number of shots). As standard procedure, we extract the action segments from the ground truth annotations to build the support set. The rest of the videos from these classes are used as query examples with this episodic benchmark (namely, using the same support set). To get reliable test results, we randomly sample 100 different episodes, creating over 1500 query samples per class. Then we compute multi-class average precision for each episodic task, and average over all episodes. The detection performance is measured via the standard mean average precision (mAP). The evaluation of our approach and SOTA are reported in Table \ref{table: untrimmed results}. The results show that TAEN can provide high results also for the temporal detection task. While we achieve comparable results to the previous work in \cite{Yang18} in 1-shot, we outperform \cite{Yang18} in 5-shots by a large margin. This shows again the higher performance expected when larger samples yield more reliable trajectories.

\begin{table}
\centering
\begin{tabular}{l c c }
Method  & mAP@0.5 & Avg. mAP \\ \hline
SMN@1 \cite{Yang18} & 22.3 & 9.8 \\
SMN@5 \cite{Yang18} & 23.1  & 10  \\
TAEN@1(I3D) & 21.99 & 9.78 \\
TAEN@5(I3D)  & 33.64 & 17.39 \\
\hline
\end{tabular}
\caption{Few-shot temporal action detection results on ActivityNet1.2
(in percentage). mAP @ tIoU threshold=$0.5$ and average mAP for the range $[0.5,0.95]$ are reported. @1 and @5 denote “1-shot” and “5-shot” respectively. }
\label{table: untrimmed results}
\end{table}
\section{Summary and future directions}
In this work, we suggest a novel idea of representing actions as trajectories in a learned feature space for few shot action recognition.
In this model, actions are encoded as trajectories in a metric space (in opposed to pixel-space) by a collection of temporally ordered sub-actions. The proposed architecture and loss function learn the coarse sub-action connectivity, by jointly learning the representation of the sub-actions per-class and the embedding function. The associated loss function optimizes for sub-action affiliation and motion between consecutive sub-actions, in the deep feature space.  For effective {\it few-shot} class discrimination we suggest a trajectory distance that combines the affiliation and motion in the feature space. The proposed network requires no additional resources or fine-tuning on the target. We further extend our model to few-shot action detection in untrimmed videos.
A recent study of \cite{TemporalAlignment} shows that using sub-action alignment improves the results. We argue that our method is more efficient by finding the optimal prototypes without the Dynamic Time Warping alignment with $\mathcal{O}(N^2)$ cost. Our method, requiring learning of only few FC layers, runs in average at rate of 6.1 $[msec/video]$ classification (excluding feature computation), on Tesla K-40, 12G.

Combining our proposed approach with fine-tuning on the few support examples of the novel categories is a good orthogonal direction that could be interesting to explore in a follow up work.

\section*{Acknowledgement}
We want to thank Eli Schwartz and Leonind Karlinsky for providing us the RepMet repo.

{\small
\bibliographystyle{ieee_fullname}
\bibliography{fewShot}

\begin{thebibliography}{10}\itemsep=-1pt

\bibitem{TARN_Bishay19}
Mina Bishay, Georgios Zoumpourlis, and Ioannis Patras.
\newblock Tarn: Temporal attentive relation network for few-shot and zero-shot
  action recognition.
\newblock {\em BMVC}, 2019.

\bibitem{Strongly_ActionLocalization_FRCNN_CVPR18}
Yu-Wei Chao, Sudheendra Vijayanarasimhan, Bryan Seybold, David~A. Ross, Jia
  Deng, and Rahul Sukthankar.
\newblock Rethinking the faster r-cnn architecture for temporal action
  localization.
\newblock {\em CVPR}, 2018.

\bibitem{ProtoGAN19}
S.~K. Dwivedi, V. Gupta, R. Mitra, S. Ahmed, and A. Jain.
\newblock {ProtoGAN:} towards few shot learning for action recognition.
\newblock {\em ICCVW}, 2019.

\bibitem{MetaLearning_fewShot_2017}
C. Finn, P. Abbeel, and S.~Levine. Model-Agnostic.
\newblock Meta-learning for fast adaptation of deep networks.
\newblock In {\em arXiv:1703.03400}, 2017.

\bibitem{Girdhar17}
Rohit Girdhar, Deva Ramanan, Abhinav Gupta, Josef Sivic, and Bryan Russell.
\newblock Actionvlad: Learning spatio-temporal aggregation for action
  classification.
\newblock {\em CVPR}, 2017.

\bibitem{Synthesis_fewShot_ICCV_2017}
B. Hariharan and R. Girshick.
\newblock Low-shot visual recognition by shrinking and hallucinating features.
\newblock In {\em ICCV}, 2017.

\bibitem{Hariharan_FewShotImageClassif17}
B. Hariharan and R. Girshick.
\newblock Low-shot visual recognition by shrinking and hallucinating features.
\newblock {\em ICCV}, 2017.

\bibitem{ActivityNet1_2_CVPR15}
B.~G. Fabian~Caba Heilbron, Victor Escorcia, and J.~C. Niebles.
\newblock A large-scale video benchmark for human activity understanding.
\newblock In {\em CVPR}, 2015.

\bibitem{Hou17}
Rui Hou, Rahul Sukthankar, and Mubarak Shah.
\newblock Real-time temporal action localization in untrimmed videos by
  sub-action discovery.
\newblock {\em BMVC}, 2017.

\bibitem{TemporalAlignment}
K.C. Jingwei, Z. Cao, C. Chang, and J. Niebles.
\newblock Few-shot video classification via temporal alignment.
\newblock {\em CVPR}, 2020.

\bibitem{FewShotObjectDetection_ICCV19}
Bingyi Kang, Zhuang Liu, Xin Wang, Fisher Yu, Jiashi Feng, and Trevor Darrell.
\newblock Few-shot object detection via feature reweighting.
\newblock In {\em ICCV}, 2019.

\bibitem{karlinsky2019repmet}
Leonid Karlinsky, Joseph Shtok, Sivan Harary, Eli Schwartz, Amit Aides, Rogerio
  Feris, Raja Giryes, and Alex~M Bronstein.
\newblock Repmet: Representative-based metric learning for classification and
  few-shot object detection.
\newblock {\em CVPR}, 2019.

\bibitem{Sports-1M_CVPR14}
Andrej Karpathy, George Toderici, Sanketh Shetty, Thomas Leung, Rahul
  Sukthankar, and Li Fei-Fei.
\newblock Large-scale video classification with convolutional neural networks.
\newblock In {\em CVPR}, 2014.

\bibitem{Kinetics17}
Will Kay, Joao Carreira, Karen Simonyan, Brian Zhang, Chloe Hillier, Sudheendra
  Vijayanarasimhan, Fabio Viola, Tim Green, Trevor Back, and Paul Natsev.
\newblock The {K}inetics human action video dataset.
\newblock {\em arXiv preprint arXiv:1705.06950}, 2017.

\bibitem{kay2017kinetics}
Will Kay, Joao Carreira, Karen Simonyan, Brian Zhang, Chloe Hillier, Sudheendra
  Vijayanarasimhan, Fabio Viola, Tim Green, Trevor Back, Paul Natsev, et~al.
\newblock The kinetics human action video dataset.
\newblock {\em arXiv:1705.06950}, 2017.

\bibitem{Kwonjoon_FewShotImageClassif19}
Kwonjoon Lee, Subhransu Maji, Avinash Ravichandran, and Stefano Soatto.
\newblock Meta-learning with differentiable convex optimization.
\newblock {\em CVPR}, 2019.

\bibitem{Lev16}
Guy Lev, Gil Sadeh, Benjamin Klein, and Lior Wolf.
\newblock Rnn fisher vectors for action recognition and image annotation.
\newblock {\em ECCV}, 2016.

\bibitem{MetricLearning_fewShot_Local_Descriptoy_CVPR_2019}
Wenbin Li, Lei Wang, Jinglin Xu, Jing Huo, Yang Gao, and Jiebo Luo.
\newblock Revisiting local descriptor based image-to-class measure for few-shot
  learning.
\newblock In {\em CVPR}, 2019.

\bibitem{lin2019bmn}
Tianwei Lin, Xiao Liu, Xin Li, Errui Ding, and Shilei Wen.
\newblock {BMN: Boundary-Matching Network for Temporal Action Proposal
  Generation}.
\newblock In {\em Proceedings of the IEEE International Conference on Computer
  Vision}, pages 3889--3898, 2019.

\bibitem{BSN_ECCV2018}
Tianwei Lin, Xu Zhao, Haisheng Su, Chongjing Wang, and Ming Yang.
\newblock {BSN: Boundary Sensitive Network for Temporal Action Proposal
  Generation}.
\newblock In {\em ECCV}, 2018.

\bibitem{Liu19}
Daochang Liu, Tingting Jiang, and Yizhou Wang.
\newblock Completeness modeling and context separation for weakly supervised
  temporal action localization.
\newblock {\em CVPR}, 2019.

\bibitem{MetaLearning_fewShot_TemporalConv_2017}
N. Mishra, M. Rohaninejad, X. Chen, and P. Abbeel.
\newblock Meta-learning with temporal convolutions.
\newblock In {\em arXiv:1707.03141}, 2017.

\bibitem{Weakly_ActionLocalization_Nguyen_CVPR18}
Phuc Nguyen, Ting Liu, Gautam Prasad, and Bohyung Han.
\newblock Weakly supervised action localization by sparse temporal pooling
  network.
\newblock {\em ECCV}, 2018.

\bibitem{W-TALC_ECCV18}
Sujoy Paul, Sourya Roy, and Amit~K Roy-Chowdhury.
\newblock {W-TALC:} weakly-supervised temporal activity localization and
  classification.
\newblock {\em ECCV}, 2018.

\bibitem{Synthesis_fewShot_NIPS_2018}
Eli Schwartz, Leonid Karlinsky, Joseph Shtok, Sivan Harary, Matias Marder,
  Abhishek Kumar, Rogerio Feris, Raja Giryes, and Alex~M. Bronstein.
\newblock {Delta-Encoder}: an effective sample synthesis method for few-shot
  object recognition.
\newblock In {\em NIPS}, 2018.

\bibitem{Two-Stream-Networks}
Karen Simonyan and Andrew Zisserman.
\newblock Two-stream convolutional networks for action recognition in videos.
\newblock In {\em NeurIPS}, 2014.

\bibitem{Snell_FewShotImageClassif17}
J. Snell, K. Swersky, and R.~S. Zemel.
\newblock Prototypical networks for few-shot learning.
\newblock {\em NeurIPS}, 2017.

\bibitem{Sung_FewShotImageClassif18}
Flood Sung, Yongxin Yang, Li Zhang, Tao Xiang, Philip~H.S. Torr, and Timothy~M.
  Hospedales.
\newblock Learning to compare: Relation network for few-shot learning.
\newblock {\em CVPR}, 2018.

\bibitem{C3D}
Du Tran, Lubomir Bourdev, Rob Fergus, Lorenzo Torresani, and Manohar Paluri.
\newblock Learning spatiotemporal features with 3d convolutional networks.
\newblock In {\em ICCV}, 2015.

\bibitem{Dense-Trajectories}
Heng Wang, Alexander Klaser, Cordelia Schmid, and Cheng-Lin Liu.
\newblock Dense trajectories and motion boundary descriptors for action
  recognition.
\newblock In {\em International Journal of Computer Vision}, 2013.

\bibitem{MetricLearning_fewShot_CVPR_2019}
Davis Wertheimer and Bharath Hariharan.
\newblock Few-shot learning with localization in realistic settings.
\newblock In {\em CVPR}, 2019.

\bibitem{FSL_ECCVW2020}
Y. Xian, B. Korbar, M. Douze, B. Schiele, Z. Akata, and L. Torresani.
\newblock Generalized many-way few-shot video classification.
\newblock {\em ECCV Workshops}, 2020.

\bibitem{R_C3D_Network_Action_Detection_ICCV17}
Huijuan Xu, Abir Das, and Kate Saenko.
\newblock {R-C3D: Region Convolutional 3D Network for Temporal Activity
  Detection}.
\newblock {\em ICCV}, 2017.

\bibitem{RC3D-arXiv2018}
H. Xu, B. Kang, X. Sun, J. Feng, K. Saenko, and T. Darrel.
\newblock Similarity r-c3d for few-shot temporal activity detection.
\newblock {\em arxiv.org/pdf/1812.10000}, 2018.

\bibitem{Yang18}
Hongtao Yang, Xuming He, and Fatih Porikli.
\newblock One-shot action localization by learning sequence matching network.
\newblock {\em CVPR}, 2018.

\bibitem{Strongly_ActionLocalization_SSN_ICCV17}
Yue Zhao, Yuanjun Xiong, Limin Wang, Zhirong Wu, Xiaoou Tang, and Dahua Lin.
\newblock Temporal action detection with structured segment networks.
\newblock {\em ICCV}, 2017.

\bibitem{Linchao18}
Linchao Zhu and Yi Yang.
\newblock Compound memory networks for few-shot video classification.
\newblock {\em ECCV}, 2018.

\bibitem{I3D}
Joao Carreira ;~Andrew Zisserman.
\newblock Quo vadis, action recognition? a new model and the kinetics dataset.
\newblock In {\em CVPR}, 2017.

\end{thebibliography}
}

\end{document}